\documentclass[a4paper, 10pt, conference]{ieeeconf}      

\IEEEoverridecommandlockouts                              

\usepackage{siunitx}
\usepackage{graphics} 
\usepackage{epsfig} 
\usepackage{mathptmx} 
\usepackage{times} 
\usepackage{amsmath}
\usepackage{amssymb}
\usepackage{mathtools}

\usepackage{url}

\usepackage{cite}

\usepackage{array}

\usepackage[caption=false,font=footnotesize]{subfig}

\makeatletter
\let\NAT@parse\undefined
\makeatother
\usepackage{hyperref}
\hypersetup{
    colorlinks=true,
    linkcolor=blue,
    filecolor=magenta,      
    urlcolor=cyan,
    pdftitle={Overleaf Example},
    pdfpagemode=FullScreen,
}

\title{\LARGE \bf
Soft Robotic Link with Controllable Transparency for \\ Vision-based Tactile and Proximity Sensing
}

\author{Quan Khanh Luu$^{1}$,
        Dinh Quang Nguyen$^{2}$,
        Nhan Huu Nguyen$^{1}$,
        and Van Anh Ho$^{1, 3*}$, \emph{Senior Member, IEEE}
\thanks{This work was supported by JST Precursory Research for Embryonic Science and Technology PRESTO under Grant JPMJPR2038.}
\thanks{$^{1}$Japan Advanced Institute of Science and Technology (JAIST), Nomi, Ishikawa, 923-1292 Japan.}
\thanks{$^{2}$Hanoi University of Industry, No.298, Cau Dien Street, Bac Tu Liem District, Hanoi, Vietnam}
\thanks{$^{3}$Japan Science and Technology Agency, PRESTO, Kawaguchi, Saitama 332-0012 Japan.}
\thanks{$^{*}$Corresponding author. Email: {\tt\small  van-ho@jaist.ac.jp}}
}

\begin{document}
\maketitle

\thispagestyle{empty}
\pagestyle{empty}

\begin{abstract}
Robots have been brought to work close to humans in many scenarios. For coexistence and collaboration, robots should be safe and pleasant for humans to interact with. To this end, the robots could be both physically soft with multimodal sensing/perception, so that the robots could have better awareness of the surrounding environment, as well as to respond properly to humans' action/intention. This paper introduces a novel soft robotic link, named \emph{ProTac}, that possesses multiple sensing modes: tactile and proximity sensing, based on computer vision and a functional material. These modalities come from a layered structure of a soft transparent silicon skin, a polymer dispersed liquid crystal (PDLC) film, and reflective markers. Here, the PDLC film can switch actively between the opaque and the transparent state, from which the tactile sensing and proximity sensing can be obtained by using cameras solely built inside the ProTac link. In this paper, inference algorithms for tactile proximity perception are introduced. Evaluation results of two sensing modalities demonstrated that, with a simple activation strategy, ProTac link could effectively perceive useful information from both approaching and in-contact obstacles. The proposed sensing device is expected to bring in ultimate solutions for design of robots with softness, whole-body and multimodal sensing, and safety control strategies.

\end{abstract}


\IEEEpeerreviewmaketitle
\section{Introduction} \label{sec: 1}

Nowadays, human-friendly robots are required to operate out of the safety fence zone, collaborating with humans in complex or physically demanding work, as in the so-called human-robot interaction (HRI) scenarios. Regarding the essential nature of these foreseen applications, the robots should be able to not only react to possible collisions, but also handle both unavoidable and intentional physical contacts in a safe and purposed manner \cite{Bicchi2008}. To potentially accomplish this, efforts should be made on the invention of novel robot components capable of passively reducing physical impacts based on compliant and lightweight structures. In addition, multimodal perception provided by embedded sensing systems permits reliable recognition of surroundings contact as well as non-contact events, which benefits the development of motion planning and control regimes possible to react purposefully in the HRI scenarios \cite{Cheng2019}. In this context, a soft sensory system endowed by both tactile and proximity sensations with a large coverage is essential.

\begin{figure}[t]
\centering 
\includegraphics[width=1.7\columnwidth]{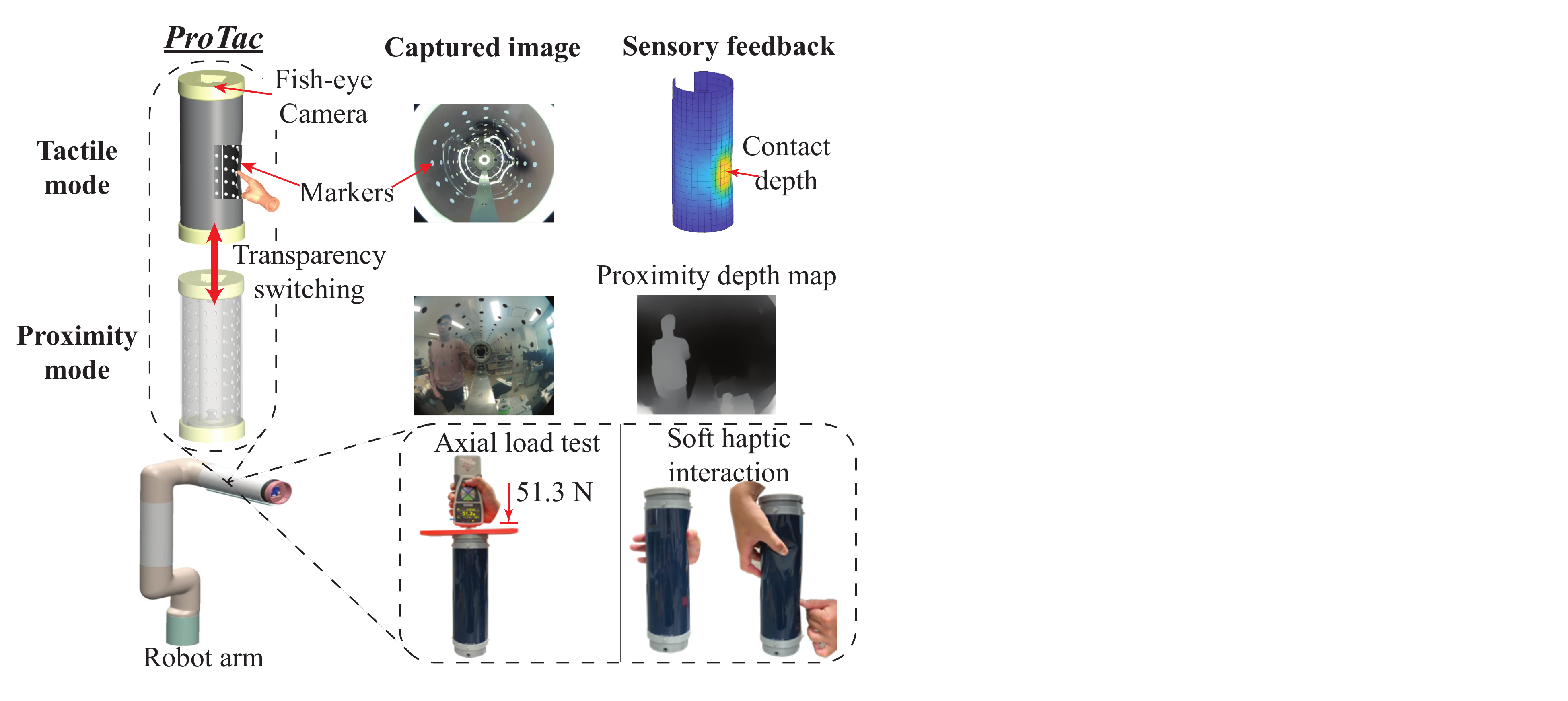}
\caption{Conceptual illustration for perceiving both tactile and proximity perception of \emph{ProTac} sensor's skin. This work is done by switching transparency of the PDLC film. \textbf{Tactile mode:} \emph{opaque} state of the skin enables cameras not to interfere with external environments, resulting in effective vision-based tactile sensing inference. \textbf{Proximity mode:} \emph{transparent} state allows the internal cameras to see through the skin for proximity perception or short-range distance measurement.}
\label{fig:paper_overview}
\end{figure}

Of sensing modalities, sense of touch offers a variety of tactile information from physical contacts, such as multiple applied forces, contact locations or complex touch patterns which are especially helpful in physical HRI (pHRI) \cite{ Li2020}. In contrast to the success of small-scaled tactile sensors, the development of tactile sensors with large coverage has faced tremendous challenges. Furthermore, most previous studies have concentrated on tactile sensing systems solely responding to physical touch and ignored touchless stimuli \cite{Navarro21}. Proximity perception has recently attracted attention for its capability of closing perception gaps induced by occlusions and blind spots in vision. One approach to combine these two sensing abilities in one sensor is attaching multiple feature-related sensing components onto a printed circuit boards encapsulated in soft material \cite{GordonCheng2011}. This system was successfully implemented in a variety of control frameworks and applications \cite{Leboutet2019, Cheng2019, Armleder2022} thanks to its conformability and scalability. However, since the sensing components are spatially distributed over the surface interspersed by electronic components, this design paradigm expressed only low/coarse spatial resolution. Not to mention the noise, since magnetic- \cite{7803315} or capacity-based \cite{5771603} tactile sensors often behaved differently with variation of material properties, which may cause difficulties in calibration and perception. 

The recent popularity of soft vision-based tactile sensors in the community is thanks to minimal wiring, high spatial resolution and low-cost \cite{s19183933}. Specifically, cameras are employed to bring in details of soft skin's deformation under physical tactile stimuli, by capturing visual features such as: 1) markers \cite{Lepora2018} and 2) reflective membrane \cite{s17122762}. Then, skin deformation will be interpreted into tactile information, including contact location, force, object texture, and so on, with help of analytical approaches \cite{Lac21} or data-based learning techniques \cite{Shotaro2021}. Sim2Real approaches \cite{Quan2022} have been recently deployed to enrich and cheapen training data sources with minimum compromise in sensing efficiency. Notably, to complement the sense of touch, a multimodal visuotactile sensor, integrated with RGB and ToF (time-of-flight) cameras, has been developed to deliver dual tactile information and proximity depth data by leveraging a selectively transmissive soft membrane \cite{Jessica2022}. Nevertheless, besides the fact that the scalability of this device to large sensing bodies is unclear, the price for the ToF camera, computation and energy cost to simultaneously process tactile and proximity images, are highly expensive.


In this paper, we introduce a novel vision-based sensing robotic link with soft skin, named as \emph{ProTac}, that possesses a large sensing area, featuring both the sense of touch at any locations on the body (\emph{i.e.,}, tactile sensing), and the sense of distance (\emph{i.e.,}, proximity) of an object/obstacle to the vicinity of the skin. Such functions are achieved by actively switching the skin optical property between \emph{opaque} (not able to be seen through) for the \emph{tactile} sensing mode and \emph{transparency} (able to be seen through) for the \emph{proximity} sensing mode (see Fig. \ref{fig:paper_overview}). In addition, we propose strategies for learning perceptions of respective sensing modalities, which can trigger the robot behavior toward safety in HRI scenarios. In summary, the contributions of this paper are shown below:

\begin{enumerate}
    \item Design and fabrication of a soft vision-based dual-mode sensing link (\emph{ProTac}), possible for selective activation of either \emph{proximity} or \emph{tactile} sensing mode based on a mechanism of \emph{switching skin transparency}. Furthermore, the same internal RGB cameras are shared for both sensing modes.
    \item Proposal of methods to learn ProTac dual-mode perception, wherein the \emph{proximity} mode measures the closest distance to external obstacles in an observed region by using monocular depth-map construction, meanwhile based on a deep learning model the \emph{tactile} mode is possible to estimate local contact depth.

\end{enumerate}


\section{Design and Fabrication} \label{sec: 2}

\subsection{Design Concept of Skin Transparency Switching}
We build ProTac upon the previous work on whole-arm tactile link (TacLink sensor) \cite{Lac21}, and extend its capability for additional proximity perception. To achieve such objective, we propose a novel combination of transparent silicone rubber and a thin flexible polymer dispersed liquid crystal (PDLC) film with attached markers. In which, the PDLC film allows switching among two modes: opaque (darken mode) and transparent mode by applying an external voltage. Here, the switching time is approximately $0.3\,$s. Therefore, the general working principle of the ProTac sensor link is as follows:
\begin{itemize}
    \item \textbf{Proximity mode}: When the PDLC film is in the transparent mode (see through), the two cameras inside the sensing link can see external objects that are close to the vicinity of the outer skin.
    \item \textbf{Tactile mode}: In contrast, the opaque (darken) mode of the PDLC film allows the two cameras to track the skin deformation through marker-featured images without external light interference, which enables the function of tactile or force sensing.
\end{itemize}

\begin{figure}[t]
      \centering
      \includegraphics[width=0.9\columnwidth]{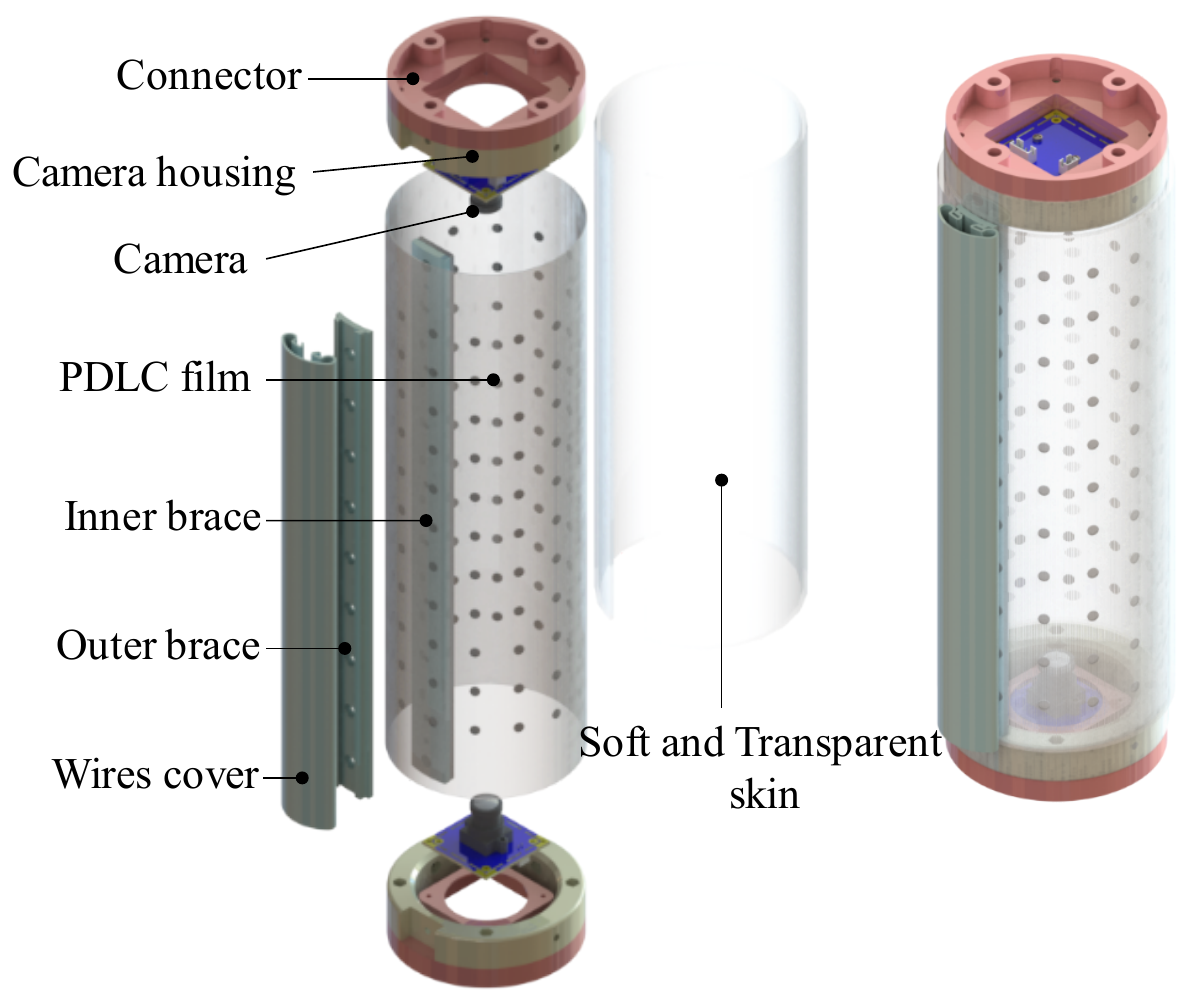}
      \caption{Design of \emph{ProTac} sensor composed of three layers skin permitting to perform two sensing modes.}
      \label{fig:design_link}
\end{figure}


Figure \ref{fig:design_link} depicts the structure of the proposed \emph{ProTac} sensor. Each end has a connector and camera housing that can accommodate a fisheye lens camera (ELP-USBFHD01M-L180). A series of LEDs were arranged circularly at the inner surface of the camera's housing to improve visibility in tactile mode. In proximity mode, they will be turned off.  Besides the use for fixing cameras, the camera housing and braces also help in shaping the cylindrical skin. On the other hand, the ProTac skin is made up of three layers: markers, PDLC film, and a deformable and transparent layer. 

\subsection{Fabrication of a ProTac Link}
The ProTac skin was basically fabricated through a molding process (see \cite{Lac21} for more details). The smoothness and evenness of the skin surface are one of the most important factors affecting the performance of both sensing modes. This can be achieved by using a set of molds with a smooth surface. Hence, a commercial acrylic tube was utilized for outside mold fabrication. Other mold parts were 3D printed using a 3D printer. Reflexive tape (R25 WHI, 3M Company) was used to form markers with a diameter of $3\,$mm. Then, the PDLC film was shaped by camera housings and the braces at two ends. Finally, a certain amount of silicon rubber was poured into the mold and the final product will be achieved after one day of curing at room temperature.

\section{Perception methodology} \label{sec: 3}
\begin{figure}[t]
\centering 
\subfloat[Tactile sensing scheme]{
\includegraphics[width=0.8\columnwidth]{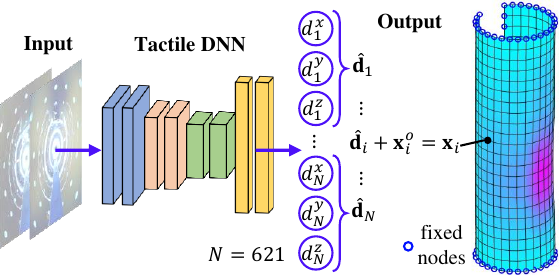}
\label{fig:tactile_sensing}
}\\

\subfloat[Proximity sensing scheme]{
\includegraphics[width=0.8\columnwidth]{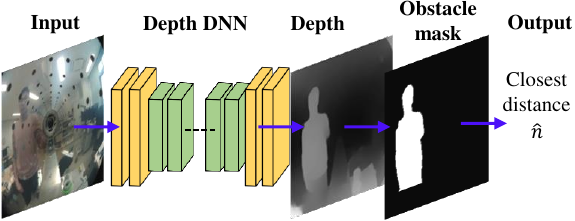}
\label{fig:proximity_sensing}
}
\caption{Explanation for sensing methodology}
\label{fig:perception_methods}
\end{figure}
Given the condition of the PDLC film (either opaque or transparent), the scheme for extracting tactile sensing and proximity sensing is shown in Fig. \ref{fig:perception_methods}. Details are as follows:
\subsection{Tactile sensing} \label{sec: tactile_sensing}
The use of model-free techniques (tactile DNN) for vision-based tactile sensing in inferring local contact depths and 3D skin shape was previously proved in \cite{Shotaro2021}. Specifically, the tactile images for each contact situation will be captured while the corresponding tactile information will be obtained from a simulation environment. In this paper, the tactile dataset was acquired by using a simulation tool based on Finite Element Method (FEM) called SOFA (Simulation Open Framework Architecture). The details of the simulation model can be referred in this work \cite{Quan2022}. Here, we assume to simulate the ProTac's skin as an isotropic, homogeneous elastomer. In more detail, the ProTac skin will be constituted by a matrix of non-overlapping tetrahedron elements (element size is $10\,$mm). Material properties follow a linear constitutive relationship (Hooke's laws) as ascribed by two parameters, Young's modulus $E = 0.22$\,\text{N/mm}$^2$ and Poisson's ratio $\nu = 0.49$. Note that $E$ was experimentally identified by a tensile test where the sample thickness is equal to the sum of soft skin and PDLC film. The simulation environment is pictured in Fig. \ref{fig:exp_setup}(a).

On the other side, an experimental setup with an identical reference coordinate system as shown in Fig. \ref{fig:exp_setup}(b) was prepared for collecting real tactile images. Then, the above dataset was used to solve a multi-output regression problem: given marker-featured tactile images $\mathbf{I}^{d}$ with the image resolution of $640\times480$ pixels; the network estimates the displacement vectors $\hat{\mathbf{D}}=[\hat{\mathbf{d}}_{i}^\top]\in\mathbb{R}^{N \times 3}$ of every free node (totally $N=621$ nodes) of the outer surface mesh representing the whole soft skin: 
\begin{equation} \label{eq:estimated_displacement_vectors}
    \hat{\mathbf{d}}_{i} \coloneqq \mathbf{x}_{i}-\mathbf{x}^{o}_{i}, \quad \forall i \in \mathbb{M},
\end{equation}
where a set $\mathbb{M}$ includes indices of free nodes; $\mathbf{x}_{i}\in\mathbb{R}^{3}$ is the 3-D position vector of one active/free node; and $\mathbf{x}^{o}_{i}\in\mathbb{R}^{3}$ is the coordinates of the respective node under the original or non-deformed state of the artificial skin. From this, the estimated local contact depth can be determined as 
\begin{equation} \label{eq:estimated_contact_depth}
    \hat{d}_{c} = \max_{i\in\mathbb{M}}||\hat{\mathbf{d}}_{i}||.
\end{equation}

The tactile DNN architecture is adapted from proven Unet convolution networks \cite{Unet}. Basically, the model consists of a contracted convolution path connected with a reverse up-convolution one via skip connections, then followed by two fully connected (FC) layers. The opaque tactile images $\mathbf{I}^{d}$ are downsampled to $256\times 256$ to establish visual inputs. Moreover, the output signal, activated by the two last FC layers, is defined by a dense single layer with $1863$ neurons to represent the estimated displacement vectors $\hat{\mathbf{D}}$, which means that we consider every $3$ adjacent neurons as a displacement vector (see Fig. \ref{fig:tactile_sensing}). For the optimization of model weights, we use iterative Stochastic Gradient Descent (SGD) optimizer with the experimentally tuned learning rate $0.015$. Details of the training loss and other network specifications can be found in \cite{Quan2022}.


\subsection{Proximity sensing} \label{sec:proximity_sensing_method}
The workflow of this method is summarized in Fig. \ref{fig:proximity_sensing} with details as below.
\subsubsection{Monocular Depth Estimation} \label{sec:depth_estimation_method}
We employed data-driven \emph{monocular} depth estimation based on a DNN to predict depth maps of external space for \emph{transparent} camera view of the ProTac link, which would form the basis for the distance measurement. In detail, we adopted the proven MiDas model \cite{midas_model} for the projection between ProTac images and estimated depth maps. The model was designed upon the ResNet multi-scale architecture \cite{He2016ResNet}, in which the input layer takes in a 3-channel ProTac image $\mathbf{I}^{t}$ (in a transparent state) and outputs an estimated depth map $\Gamma$; both have the same image resolution of $640\times480$. The model was trained on diverse existing datasets (\emph{e.g.}, ReDWeb \cite{ReDWeb}, MegaDepth \cite{Li2018_MegaDepth}, and WSVD \cite{wang2019web}) using a scale-invariant depth regression loss as mentioned in \cite{midas_model}. For the training process, the Adam optimizer was adopted with the learning rate initialized at $10^{-5}$ for the encoder path pre-trained on Imagenet \cite{ImageNet} and $10^{-4}$ for other layers, and then linearly decaying at the 50$^{\text{th}}$ iteration out of a total of 100 training steps. In order to increase the generalization, the input images with aspect ratio maintained were randomly flipped, cropped, and then resized to $384\times384$, with $50\%$ chance. Details of network architecture can be found in \cite{He2016ResNet}.

\subsubsection{Distance Detection} \label{sec:distance_detection}
Whilst depth estimation could directly provide plenty of high-level information in perception systems, this preliminary work focuses on the closest normal distance between nearby obstacles and the ProTac link based on the estimated depth map $\Gamma$. 

Toward this goal, the image mask of a target nearby obstacle ($\Xi$) is first extracted from $\Gamma$ using Otsu's automatic binary thresholding (OpenCV module), by which we assume the proximal objects of interest would have distinguishable, bright pixel intensity. The subsequent problem is to compute the normal distances from the masked obstacles to the skin surface. Given the camera was modeled as a classic pinhole, the 3D positions of the masked obstacles $\mathbf{O}=[\mathbf{o}_{k}^\top]\in\mathbb{R}^{K \times 3}$ ($K$ is the number of masked pixels) could be calculated as:
\begin{equation}
\begin{split}
    o^{x}_{k} & = \frac{(u_{k} - c_{x})(b+o^{z}_{k})}{f_{x}},\,o^{y}_{k} = \frac{(v_{k} - c_{y})(b+o^{z}_{k})}{f_{y}}, \\
    o^{z}_{k} & =\Gamma_{k}, \quad \forall k \in \mathbb{K},
\end{split}
\end{equation}
where $\mathbb{K}=\{i\,|\, \Gamma_{i}\wedge\Xi_{i} = 1\}$ is indices of masked obstacles; $\mathbf{o}_{k} = [o^{x}_{k},\,\,o^{y}_{k},\,\,o^{z}_{k}]^\top$ is the 3D position of an obstacle point; $b$ denotes the position of PCS (ProTac coordinate system) origin in the camera frame; $(u_{k}, v_{k})$ is the geometrical projection of $\mathbf{o}_{k}$ on the image plane; $f_{x}$ and $f_{y}$ are the focal lengths along $x$- and $y$-axes, respectively; $(c_{x}, c_{y})$ is the pixel location of the image principal point on the pixel coordinate $\{u, v\}$. The identification of model parameters $\{f_{x}, c_{x}, c_{y}, b\}$ and fisheye-lens correction were conducted following the method proposed in \cite{Lac21}. For ease of calculating the normal distance to the skin surface based on the location of the obstacles $\mathbf{O}$, the Cartesian coordinates of obstacle points are converted to a radial coordinate as $r_{k} = \sqrt{o^{x2}_{k} + o^{y2}_{k}}$. Thus, the normal distance between the obstacle points and the skin surface can be estimated as
\begin{equation}\label{eq:estimated_proximity_distance}
    n_{k} = r_{k} - r^{s}, \quad \forall k \in \mathbb{K},
\end{equation}
where $r^{s}$ is the radial of the sensor skin. From this, we could finally perform a quick search for the closet distance:
\begin{equation} \label{eq:estimated_closest_distance}
    \hat{n} = \min_{k\in\mathbb{K}}n_{k}.
\end{equation}

This proposed procedure for distance detection is derived for a \emph{single} camera view, while a combination of the two opposite cameras just as the current ProTac setup could provide a larger sensing range from any direction, which mitigates possible occlusions. 

\section{Performance evaluation} \label{sec: 4}
\subsection{Experimental setup} \label{sec: exp_setup}
\begin{figure}[t]
      \centering
      \includegraphics[width=0.6\columnwidth]{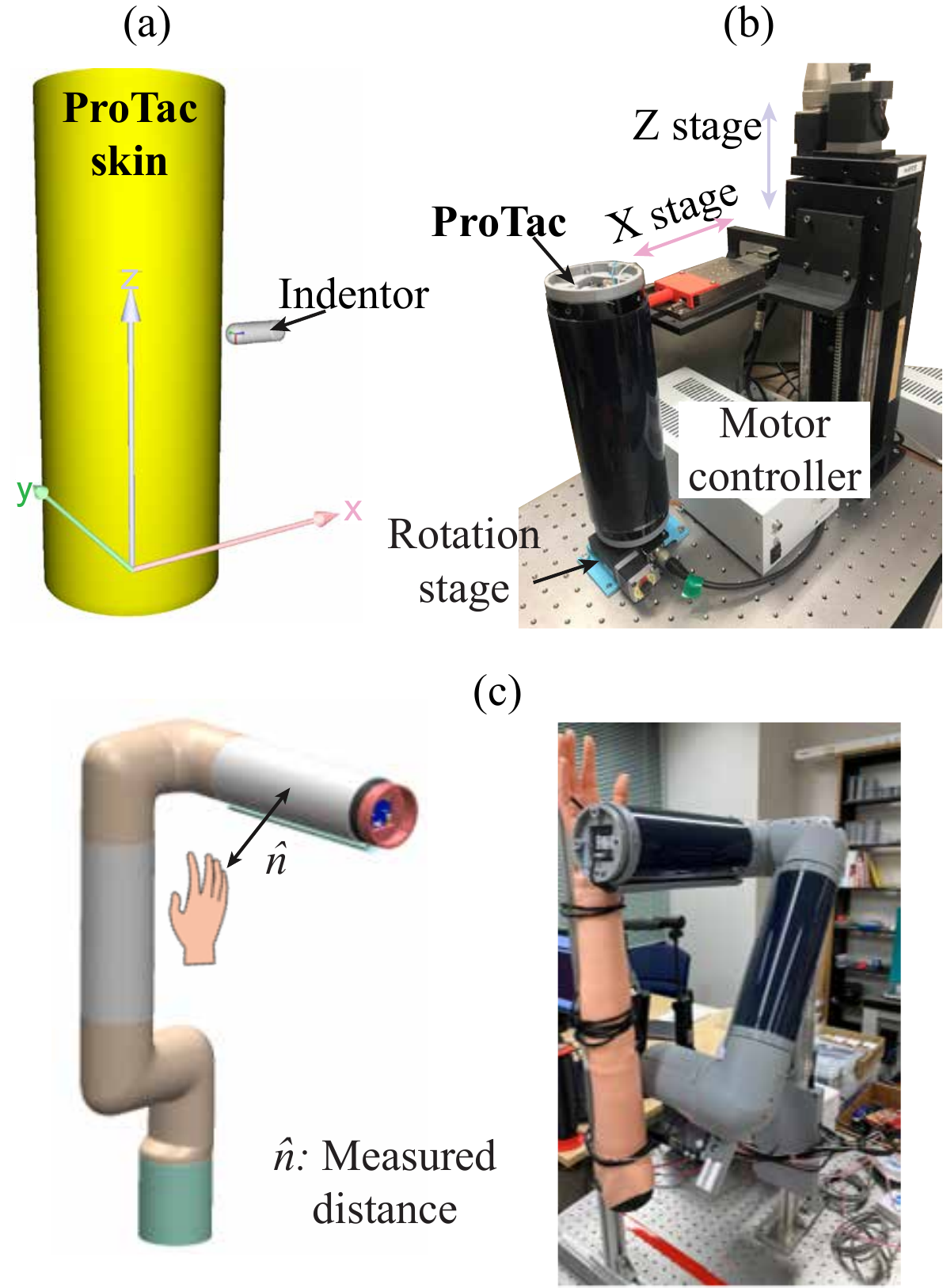}
      \caption{Illustrations of (a) simulation environment (SOFA) and (b) actual experiment scheme for training data collection. Figure (c) pictures the experiment setup for distance estimation in proximity mode.}
      \label{fig:exp_setup}
\end{figure}
\begin{figure*}[t]
\centering 
\subfloat[Measured contact depth versus true value.]{
\includegraphics[width=0.6\columnwidth]{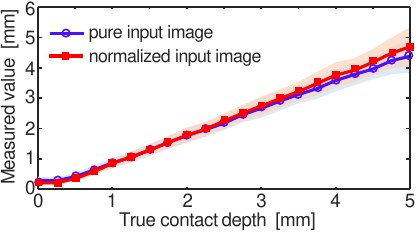}
\label{fig:contact_depth}
}
\hfill
\subfloat[Absolute measurement error with increased contact depth.]{
\includegraphics[width=0.6\columnwidth]{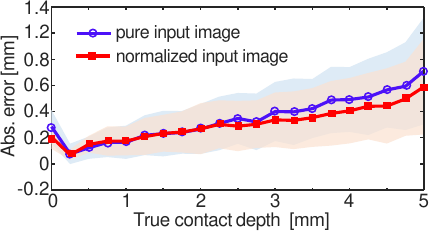}
\label{fig:abs_error}
}
\hfill
\subfloat[Estimated contact depths (ground-truth values = $2.5,\,4,\,5$\,mm) with various input images.]{
\includegraphics[width=0.6\columnwidth]{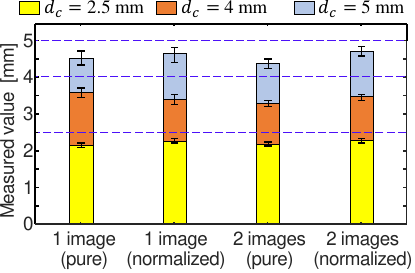}
\label{fig:input_comparision}
}\\
\caption{Evaluation of contact depth accuracy estimated based on different input signals.}
\label{fig:reconstruction_accuracy}
\end{figure*}

\subsubsection{Tactile mode} 

The testbed shown in Fig. \ref{fig:exp_setup}b was used to collect the tactile images for training the DNN model, as well as to evaluate the performance of the tactile mode. It is comprised of two linear stages, a rotating stage (Suruga Seiki Co., Japan) and a stepping motor controller (DS102, Suruga Seiki Co., Ltd., Japan). The $X$-axis stage drives a spherical-head indentor ($12\,$mm diameter) to make contact with the desired depth (maximum depth is $5\,$mm) at the exact positions of free nodes. The contact locations were secured by the horizontal movement of the indentor ($Z$-axis linear carrier) and the rotation of the ProTac sensor (rotation stage). It should be noted that the nominal axis of the indentor was pre-adjusted in advance to intersect with the $Z$-axis.

\subsubsection{Proximity mode} 
The ability to estimate ProTac-obstacle distance ($\hat{n}$) was studied with the experiment arrangement shown in Fig. \ref{fig:exp_setup}c. Specifically, an artificial human arm was vertically fixed onto a linear translation guide mechanism so that the hand directly faces a ProTac link. Then, the following procedure was conducted: 1) Put the arm slightly in touch with the skin and set this position as the origin (ground-truth distance $n = 0\,$mm), 2) Move the arm far away from the skin with the distance up to $100\,$mm in moving step of $10\,$mm. For each step, 150 samples of estimated distance Eq. \eqref{eq:estimated_closest_distance} using the algorithm in Section \ref{sec:proximity_sensing_method} were recorded to judge the sensing performance. The results of proximity sensing are presented in Section \ref{sec: result_proximity}.

\subsection{Tactile sensing mode} \label{sec: result_tactile}
For tactile sensing, we verify the accuracy of contact depth $\hat{d}_c$ estimated by the trained DNN with different types of input images. For evaluation, Unet-based DNN with $2048$ neurons for each of the last two FC layers was employed, since through our experimental trials it showed to outperform other model backbones. The model was trained with $80\%$ of total $11025$ samples, in which $20\%$ of the dataset was withheld for validation.  

The experimental results showed that measurement errors increased with true contact depth ($d_{c}$) in both cases of non-normalized (pure) and normalized visual inputs that have two tactile images concatenated (see Figs. \ref{fig:reconstruction_accuracy}a-b). However, normalized inputs yielded higher estimation accuracy as larger contact intensities, compared with the pure ones. In fact, the averaged absolute errors at $d_{c}=5\ $mm were more or less $0.7\ $mm and $0.6\ $mm, which approximate full-scale errors $14\%$ and $12\%$ (with FS $5\,$mm) for pure and normalized inputs, respectively. We also evaluated the sensing performance based on either the single-view visual input or concatenated two opposite camera views, each with and without normalization at different contact depths ($d_{c}=2.5,\,4,\,5\,$mm). The result showed that the inputs with normalization yielded better performance in both cases of single- and double-view input (see Fig. \ref{fig:input_comparision}). In addition, by concatenating two normalized tactile images, the measurement accuracy was improved by around $1.45\%$ and $1.09\%$ in terms of full-scale error at $d_{c}=4$ and $5\,$mm, respectively, as compared to the single-view images.

\begin{figure}[ht]
\centering 
\includegraphics[width=0.75\columnwidth]{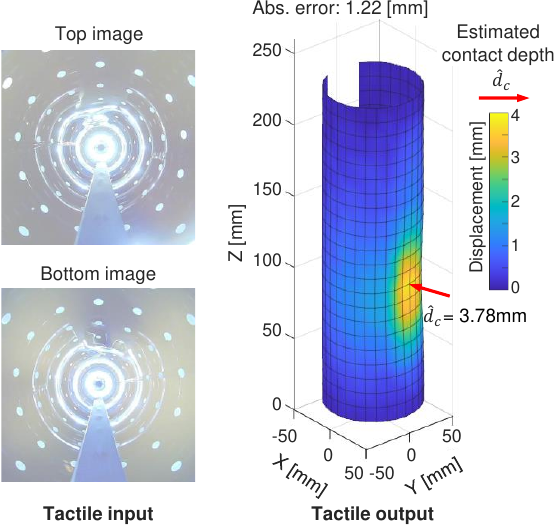}
\caption{The visualization of DNN-based 3D reconstruction and its estimated contact depth compared to ground-truth $d_{c}=5\,$mm. The model takes the two concatenated tactile images (in the dark mode) as tactile input.}
\label{fig:tactile_visulization}
\end{figure}

Lastly, we showcase the visualization of the skin shape reconstruction in the critical scenario that contact happens near the middle of the link; where the visual clue is farthest from both cameras (see Fig. \ref{fig:tactile_visulization}). In this case, with the contact depth $d_{c}=5\,$mm, the concatenated inputs caused an estimation accuracy of $1.22\,$mm and $24.4\%$ with respect to the absolute and full-scale errors, respectively. 

\textit{Discussion}: The obtained results demonstrate the normalization of tactile inputs is necessary for improved sensing performance. Moreover, since the inference accuracy based on the single-view input is rather comparable with the concatenated one, we would argue that this sensing method is still applicable for small-sized tactile sensors with a single camera while combining camera views could yield better performance for such a large-scale device like the one showcased in this paper. It is also worth noting that although it is reasonable to recognize multi-point contact depth/location from the nodal displacement vector $\hat{\mathbf{D}}$, we will elaborate it in our future work.  

\subsection{Proximity sensing mode} \label{sec: result_proximity}
\begin{figure}[ht]
\centering 
\subfloat[RGB camera view]{
\includegraphics[width=0.9\columnwidth]{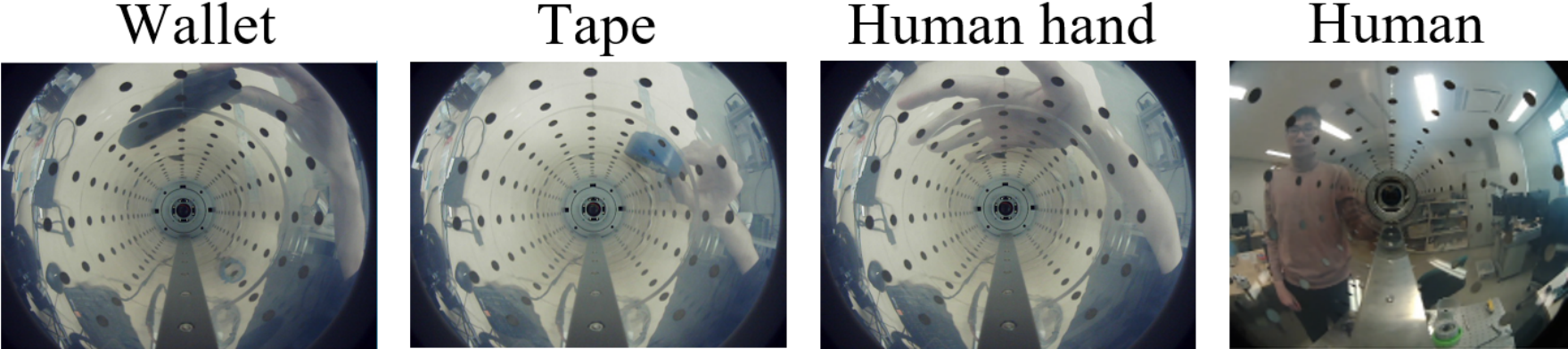}
\label{fig:rgb_view}
}\\

\subfloat[Nearby object mask]{
\includegraphics[width=0.9\columnwidth]{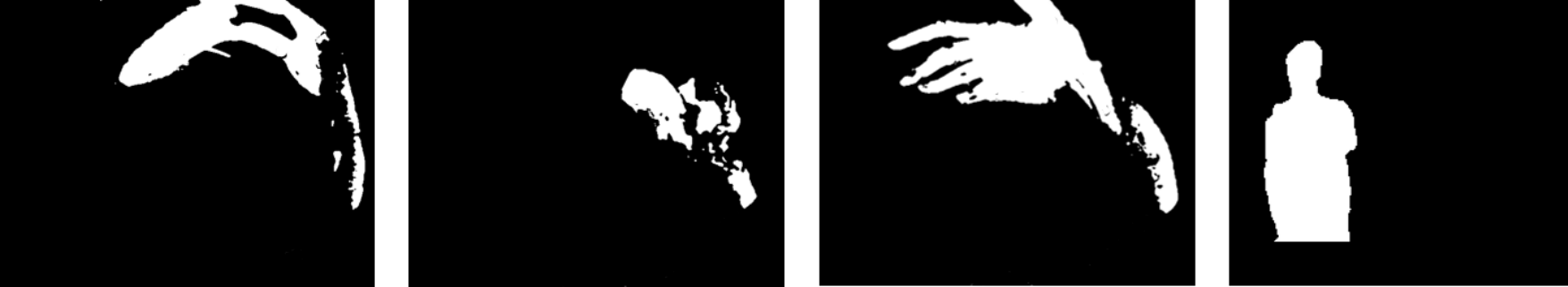}
\label{fig:obstacle_mask}
}
\caption{Examples of object mask extraction from transparent camera views.}
\label{fig:prox_vis_results}
\end{figure}

This section summarizes initial findings of the proximity-sensing abilities as the ProTac skin is in the transparent state. Figure \ref{fig:prox_vis_results} highlights the recognition of plenty of different obstacles (\emph{e.g.,} wallet, tape, hand, and human) from the corresponding transparent views (see Fig. \ref{fig:rgb_view}), which is expressed via the obstacle masks $\Xi$ (Fig. \ref{fig:obstacle_mask}).

\begin{figure}[ht]
\centering 
\subfloat[ProTac-obstacle distance measurement accuracy]{
\includegraphics[width=0.92\columnwidth]{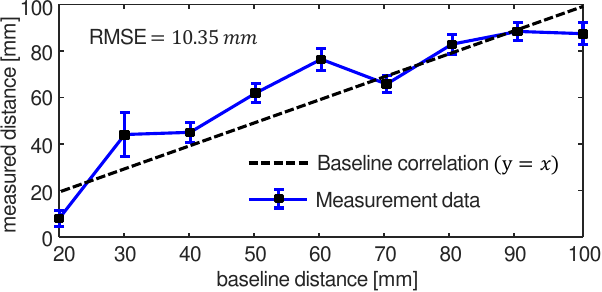}
\label{fig:distance_accuracy}
}\\

\subfloat[Distance measurements at $n=80\,mm$]{
\includegraphics[width=0.9\columnwidth]{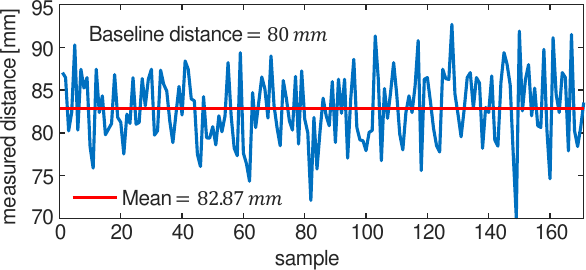}
\label{fig:measurment_samples}
}
\caption{The evaluation of ProTac-based distance sensing. The experimental setup can be found in Section \ref{sec: exp_setup}.}
\label{fig:prox_distance_measurment}
\end{figure}

The accuracy of the calibrated closest distance estimated by the ProTac link is reported in Figure \ref{fig:distance_accuracy}, in which the estimation accuracy reached approximately $10.35\,$mm in terms of RMSE (root mean squared error) metric averaged over a reliable range $n\in[20,\,100]\,$mm. Despite the fact the upper sensing range could be further than $100\,$mm, the distance out of the reliable range caused a lot of uncertainties because the DNN model for depth estimation did not yet fine-tune for specific ProTac images. Lastly, Figure \ref{fig:measurment_samples} shows the log of measurement samples at the baseline distance $n=80\,$mm, which yielded the measured distance mean $\hat{n}=82.87\,$mm.

\textit{Discussion}: These preliminary results with the provided proximity sensing method promise a wider sensing range beyond $100\,$mm once fine-tuning the depth-map model is conducted. Also, the proposed method leaves room for the combination of two camera views in order to enlarge the sensing area, as well as for determining the approaching direction of nearby obstacles, which will be thoroughly examined in our future work. 

\section{Conclusion and future work} \label{sec: 5}
This work introduces a novel design for a soft robotic link (ProTac) which features both tactile and proximity perception enabled by changing the transparency of the skin (the PDLC film in particular). The presented design and fabrication process can be extended to other vision-based tactile sensors in literature to date. Furthermore, the evaluation results of system performance in both modes certify the feasibility and potential of this idea in real robotic arms.

However, there are still some aspects that need further elaboration. Firstly, the simulation model used to collect referenced tactile data for model training does not accurately reflect the behaviors of ProTac skin. The mechanical properties of the ProTac skin should be attributed to the coupling of the silicon-made soft layer and the PDLC film. Secondly, the mechanical stability of the robot arm constituted by ProTac links in high-load operation is still questionable. An in-depth structural analysis for static and dynamic manner should be addressed in future work. Moreover, it is necessary to attain an optimal strategy for switching back and forth between two sensing modes to facilitate dexterous control and perception of the robot arm in different tasks. The effectiveness of multimodal perception in the creation of a safe and intelligent human-robot interaction should be further investigated.

\bibliographystyle{IEEEtran}
\bibliography{IEEEabrv, ref}

\end{document}